\documentclass[letterpaper]{article} 
\usepackage{aaai2026}  
\usepackage{times}  
\usepackage{helvet}  
\usepackage{courier}  
\usepackage[hyphens]{url}  
\usepackage{graphicx} 
\usepackage{natbib}  
\usepackage{caption} 
\usepackage{algorithm}
\usepackage{algorithmic}

\usepackage{newfloat}
\usepackage{listings}
\usepackage{amsmath}  
\usepackage{enumitem}
\usepackage{amssymb}

\DeclareCaptionStyle{ruled}{labelfont=normalfont,labelsep=colon,strut=off} 
\floatstyle{ruled}
\newfloat{listing}{tb}{lst}{}
\floatname{listing}{Listing}
\title{Do Large Language Models Think Like the Brain? \\ Sentence-Level Evidences from Layer-Wise Embeddings and fMRI}

\author{
    Yu Lei\textsuperscript{\rm 1}\thanks{These authors contributed equally.}\thanks{Corresponding authors.},
   Xingyang Ge\textsuperscript{\rm 2,3}\footnotemark[1],     
   Yi Zhang\textsuperscript{\rm 4,5}, 
   Yiming Yang\textsuperscript{\rm 3,6}\footnotemark[2], 
   Bolei Ma\textsuperscript{\rm 4,7}\footnotemark[2]
}
\affiliations{
    \textsuperscript{\rm 1}School of Artificial Intelligence, Beijing University of Posts and Telecommunications\\

    \textsuperscript{\rm 2}School of Literature, Shandong University \\
    \textsuperscript{\rm 3}School of Linguistic Sciences and Arts, Jiangsu Normal University\\
    \textsuperscript{\rm 4}Department of Statistics, LMU Munich\\
    \textsuperscript{\rm 5}FAU Erlangen-Nuremberg\\
    \textsuperscript{\rm 6}Collaborative Innovation Center for Language Ability, Jiangsu Normal University \\
    \textsuperscript{\rm 7}Munich Center for Machine Learning\\
    leiyu@bupt.edu.cn, 202320175@mail.sdu.edu.cn, yi.zhang@fau.de, yangym@jsnu.edu.cn, bolei.ma@lmu.de
}

\usepackage{bibentry}

\begin{document}

\maketitle

\begin{abstract}
Understanding whether large language models (LLMs) and the human brain converge on similar computational principles remains a fundamental and important question in cognitive neuroscience and AI. Do the brain-like patterns observed in LLMs emerge simply from scaling, or do they reflect deeper alignment with the architecture of human language processing? This study focuses on the sentence-level neural mechanisms of language models, systematically investigating how layer-wise representations in LLMs align with the dynamic neural responses during human sentence comprehension. By comparing hierarchical embeddings from 14 publicly available LLMs with fMRI data collected from participants, who were exposed to a naturalistic narrative story, we constructed sentence-level neural prediction models to identify the model layers most significantly correlated with brain region activations. Results show that improvements in model performance drive the evolution of representational architectures toward brain-like hierarchies, particularly achieving stronger functional and anatomical correspondence at higher semantic abstraction levels. These findings advance our understanding of the computational parallels between LLMs and the human brain, highlighting the potential of LLMs as models for human language processing.
\end{abstract}

\begin{links}
    \link{Code}{https://github.com/Lucasuuu02/LLM4Brain}
\end{links}

\section{Introduction}

The intersection of artificial intelligence and neuroscience has emerged as a cutting-edge research frontier, particularly in understanding the parallels between large language models (LLMs) and human neural language processing~\cite{toneva2019interpreting, liu2025outraged, schrimpf2021neural}. Prior studies~\cite{anderson2021deep,caucheteux2021disentangling} have established evidence showing intriguing correlations between LLM-learned representations and neural responses during language processing~\cite{zhao2025agentcdm}, particularly in feature extraction and representational similarity~\cite{caucheteux2022brains,hosseini2022artificial}. These findings~\cite{sun2020neural} suggest that both systems may utilize comparable linguistic features, as evidenced by the linear mappability of LLM representations to neural activity patterns. However, these observations lack mechanistic explanations for what critical properties enable LLMs to achieve brain-like processing capabilities.
\begin{figure}[t]
    \centering
    \includegraphics[width=0.8\linewidth]{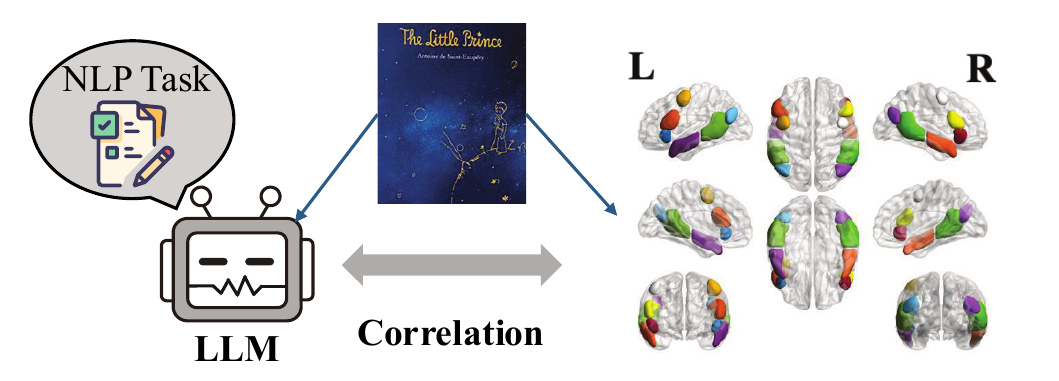}
    \caption{A brief presentation of the experimental design. We expose both LLMs and real humans to a narrative story (``The Little Prince'') and aim to compare the correlation between LLMs' and human brains' language processing.}
    \label{fig:brain}
\end{figure}

Recent investigations~\cite{goldstein2022shared,caucheteux2023evidence} have explored multimodal similarities between LLMs and neural processes. While some studies~\cite{antonello2023scaling,antonello2024predictive}  demonstrate stronger alignment between autoregressive LLMs~\cite{ethayarajh2019contextual,tenney2019bert,lei2024fairmindsim} and the predictive coding hypothesis of human language processing, others focus on metrics like language modeling performance, model scale, and representational generalization~\cite{liu2025prosocial}. Collectively, these works indicates that modeling quality critically determines brain-like representational capacities \cite{hickok2007cortical,lerner2011topographic}. A fundamental question remains unresolved: \emph{Does this similarity merely stem from increased model scale, or does it reflect deeper convergence in computational principles with the human speech processing pathway?} Resolving this dichotomy is crucial for advancing next-generation model architectures.

Current studies examining the relationship between LLMs and brain decoding typically use publicly available datasets to evaluate model performance in open scenarios. However, such datasets may fail to accurately reflect the models' comprehension abilities in specific tasks. To address this limitation, we selected 14 publicly available pre-trained LLMs and designed a sentence understanding task to assess their contextual understanding using the same text materials used for human fMRI. Additionally, we utilized fMRI data from participants who listened to naturalistic text to construct decoding models, using the correlation metric to compare the relationship between LLMs and brain-activity patterns in relevant regions. A simple presentation of our experimental design is illustrated in Figure \ref{fig:brain}.

We summarize our main findings as follows:

\begin{enumerate}[leftmargin=*,nolistsep]
\item[1)] \textbf{Instruction tuning boosts performance and brain alignment:} Instruction-tuned models consistently outperformed their base versions in both sentence comprehension and neural alignment, demonstrating statistically significant improvements.
\item[2)] \textbf{Intermediate layers better reflect brain activity:} All LLMs showed stronger correlations with brain responses at intermediate layers, and models with higher comprehension ability exhibited stronger neural alignment.
\item[3)] \textbf{Hemispheric asymmetry reveals functional specialization:} Left-dominant regions aligned with core language processing, while right-hemisphere regions reflected higher-level cognitive functions, both shaping model–brain correspondence.
\end{enumerate}


\section{Related Work}

\paragraph{Neuroscientific Foundations of Language Comprehension.}
Before the advent of the current large-scale 
LLMs, cognitive neuroscience had established that human language comprehension relies on complex hierarchical processing~\cite{friederici2011brain}. Research centered on understanding how the brain integrates perceptual units (e.g., phonemes and graphemes) into meaningful structures at the lexical, sentential, and discourse levels~\cite{price2012review}. Early models, such as the Wernicke-Lichtheim-Geschwind model~\cite{geschwind1967wernicke}, identified localized brain regions for language but struggled to explain sentence-level and discourse-level processing, especially information integration across sentences for coherence~\cite{hickok2004dorsal}. Cognitive Neuroscience Methods like fMRI and ERP~\cite{kutas1984brain, osterhout1992event} led to a shift from localized to distributed network models, such as the Memory-Unification-Control (MUC) framework~\cite{hagoort2016muc}. However, early fMRI studies were limited by isolated stimulus presentation and signal averaging, which made capturing long-range linguistic integration difficult~\cite{humphries2007time}. New naturalistic paradigms like narrative listening~\cite{brennan2016naturalistic} advanced investigation into discourse comprehension, but approaches like representational similarity analysis (RSA)~\cite{kriegeskorte2008representational} often underestimate discourse features like coherence and context-aware meaning~\cite{zacks2017cognitive, messi2025tracking}, emphasizing the need for multi-sentence integration analyses.

\paragraph{Leveraging LLMs for Brain-Language Mapping.}
Recent studies~\cite{luo2022cogtaskonomy, yu2024predicting, fu2025headsbetteronedistilling} use the strong semantic capabilities of pretrained language models to examine brain-language mappings and decode neural processes. For instance,~\citet{ren2024large, yu2024predicting} applied Dynamic Similarity Analysis (DSA) to compare text embeddings with fMRI signals, constructing Representational Dissimilarity Matrices (RDMs) using measures like Pearson correlation. Other works~\cite{mischler2024contextual, bonnasse2024fmri} aligned layer-wise activations of language models with averaged fMRI activity maps using ridge regression. Moreover,~\citet{tuckute2024driving} trained encoding models on fMRI data from participants exposed to diverse sentences, optimizing GPT-2 XL embeddings for neural alignment. There is also evidence that high-level visual representations in the human brain are aligned with LLMs \cite{Doerig2025}. These studies showcase the potential of LLMs to reveal insights into the neural mechanisms of language comprehension, offering new tools and methodologies for cognitive neuroscience.

\begin{figure*}[ht]
    \centering
    \includegraphics[width=0.75\linewidth]{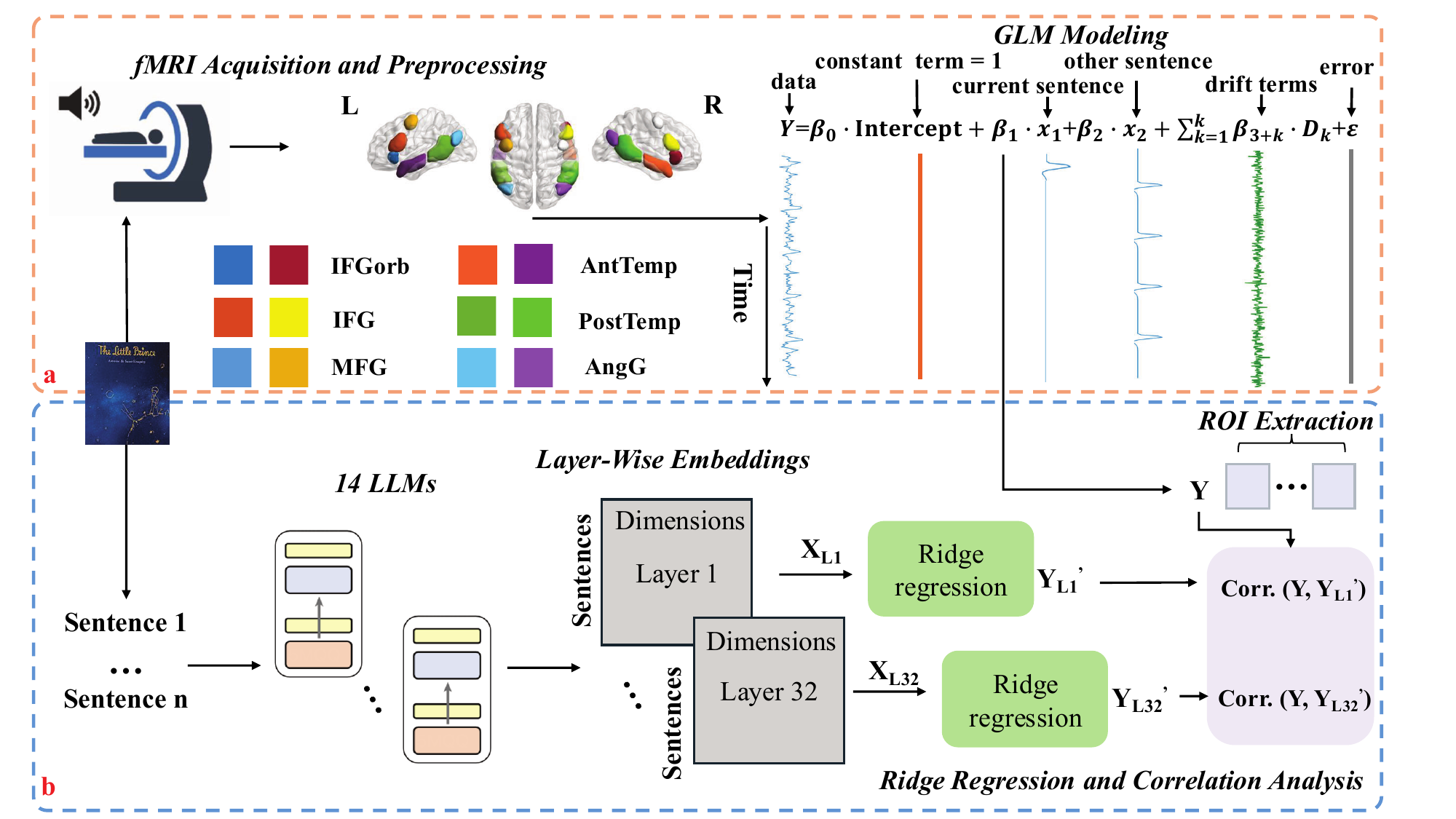}
    \caption{
    Multi-stage pipeline to analyze the alignment between LLM representations and neural responses during naturalistic language comprehension. The methodology includes auditory stimulus presentation, layer-wise embedding extraction from LLMs, voxel-wise regression modeling, and region-of-interest (ROI)-based brain-model alignment analysis. Panel (a) outlines the neuroimaging data acquisition and preprocessing steps, while panel (b) describes the brain-LLM alignment analysis.}
    \label{fig:method}
\end{figure*}

\section{Methodology}
To examine the alignment between LLM representations and brain activity during naturalistic language comprehension, we implement a multi-stage pipeline (Figure~\ref{fig:method}). We first preprocess the fMRI data and use a General Linear Model (GLM) to estimate sentence-level neural responses. Language-related ROIs are then extracted for analysis. We compute sentence embeddings from different LLM layers and evaluate their correspondence with brain activity using voxel-wise encoding models.



\subsection{Data and Stimuli}
\label{sec:participants}
We used the existing data from \citet{Li2022} which provides the ``The Little Prince'' multilingual naturalistic fMRI corpus. The dataset included 35 healthy, right-handed, native Mandarin speakers (15 female; mean age 19.3, range 18–25), but one participant was excluded due to incomplete data. The stimuli used are parallel corpora of the diary ``The Little Prince'' in both English and Chinese. The texts are segmented into 1,577 sentences in Chinese, with corresponding English sentences aligned for each one.


\subsection{fMRI Acquisition and Preprocessing}
\label{sec:acquisition}
Stimuli were presented via PsychoPy 2 and MRI-compatible headphones. Participants listened to a 99‑min Chinese audiobook (nine 10‑min segments) and answered four questions after each (36 total) via head‑coil mirror and button box. Total session ~2.5 h.

Data acquisition used a 3T GE Discovery MR750 scanner (32-channel coil). Structural images: T1-weighted MP-RAGE sequence; Functional images: multi-echo planar imaging (ME-EPI) sequence (TR = 2000 ms; TEs = 12.8, 27.5, 43 ms; voxel size = 3.75 × 3.75 × 3.8 mm). Preprocessing was performed using AFNI 16 (Cox, 1996), including removal of the first 4 time points, multi-echo independent component analysis (ME-ICA) for denoising, and spatial normalization to the MNI standard space (resampled to 2 mm³ voxels).

\subsection{General Linear Model (GLM)}  
\label{sec:glm}  

The general linear model (GLM) serves as the standard analytical framework in fMRI neuroscience research, employing statistical modeling to accurately isolate task-evoked Blood oxygenation level dependent (BOLD) signals from physiological and system noise. The core model is: 

{\small
\begin{equation}
  \mathbf{Y} = \mathbf{X}\boldsymbol{\beta} + \boldsymbol{\epsilon}, \quad \boldsymbol{\epsilon} \sim \mathcal{N}(\mathbf{0}, \sigma^2\mathbf{I})
\end{equation}
}
where $\mathbf{Y}$ represents continuous BOLD time series sampled at each TR, $\mathbf{X}$ is the design matrix encoding experimental conditions and nuisances, $\boldsymbol{\beta}$ contains regression coefficients, and $\boldsymbol{\epsilon}$ is residual noise.

The design matrix $\mathbf{X}$ is constructed by convolving event onsets with a canonical hemodynamic response function (HRF).Crucially, this convolution process uses the precise onset time of each sentence, not the TR, thereby resolving the temporal mismatch between sentence boundaries and the discrete TR sampling intervals. Each condition's regressor $\mathbf{x}_j$ includes the trial onsets and nuisance confounds (e.g., motion, drift):

{\small
\begin{equation}
  \mathbf{x}_j(t) = \sum_{k=1}^{K_j} \text{HRF}(t - t_{\text{onset}}^{(j,k)}) + \sum_{m=1}^M \gamma_m \mathbf{n}_m(t)
\end{equation}
}

We adopt the Least-Squares Separate (LS-S) approach~\cite{mumford2014impact} to isolate BOLD responses for each sentence-level trial by modeling each target sentence independently. This method reduces collinearity and improves sensitivity to transient, event-related activity compared to block-based or condition-averaged models. LS-S enables precise sentence-level neural activation patterns under naturalistic stimuli.

Voxel-wise parameter estimates are obtained as:

{\small\begin{equation}
  \hat{\boldsymbol{\beta}} = (\mathbf{X}^\top \mathbf{X})^{-1} \mathbf{X}^\top \mathbf{Y}
\end{equation}
}
$t$-statistics are computed for testing contrasts $\mathbf{c}$:

{\small
\begin{equation}
  t_v = \frac{\mathbf{c}^\top \hat{\boldsymbol{\beta}}_v}{\sqrt{\hat{\sigma}_v^2 \cdot \mathbf{c}^\top (\mathbf{X}^\top \mathbf{X})^{-1} \mathbf{c}}}, \quad \hat{\sigma}_v^2 = \frac{\|\mathbf{Y}_v - \mathbf{X}\hat{\boldsymbol{\beta}}_v\|^2}{T - \text{rank}(\mathbf{X})}
\end{equation}
}

This enables inference on condition-specific activations while accounting for noise and drift, improving sensitivity and interpretability of neural responses to linguistic stimuli.

\subsection{Region of Interest (ROI) Extraction}
\label{sec:roi}

We then extract the ROIs using the language network proposed by~\cite{fedorenko2010new}, which serves as the core analytical framework for our investigation into the neurobiological distinctions between human brains and LLMs during natural language comprehension. This network has been widely validated as the neural substrate for representing linguistic knowledge and supporting key language processes \cite{tuckute2024driving}. 

\subsection{Layer-Wise Embeddings of Sentences}
\label{sec:sentence_embedding}
We then introduce the LLM processing of ``The Little Prince'', as shown in Figure \ref{fig:method}(b). The text is segmented into sentences and then fed into the LLMs. 

We extract layer-wise embeddings from 14 language models of varying depths. Each sentence is processed through model-specific sliding windows, capturing all hidden states at terminal positions. Segment representations are averaged into layer × dimension tensors, preserving native dimensionality per model for neural decoding.

\subsection{Ridge Regression and Correlation Analysis}
\label{sec:Cross-Validated}  

In this framework, we employ ridge regression to quantify how semantic representations from different layers of a neural network relate to brain activity patterns captured via fMRI. By integrating deep learning and neuroscience, we systematically evaluate how well neural embeddings predict fMRI signals for a specific region of interest (ROI). Given fMRI response vectors $\mathbf{y} \in \mathbb{R}^N$ and neural embeddings $\mathbf{X} \in \mathbb{R}^{N \times L \times D}$ (where $T$ is the number of time points, $L$ is the number of layers, and $D$ denotes embedding dimensionality), the predictive performance for each layer is computed as the average Pearson correlation ($\rho$) across $K$ cross-validation folds:

{\small
\begin{equation}
  \rho_l = \frac{1}{K} \sum_{k=1}^K \text{corr}\left( \mathbf{y}_{\text{test}}^{(k)}, \mathbf{X}_{\text{test}}^{(l,k)} \hat{\boldsymbol{\beta}}^{(l,k)} \right)
\end{equation}
}

The model uses ridge regression to estimate these predictions, balancing data fidelity and regularization to avoid overfitting. Regression weights ($\boldsymbol{\beta}$) are optimized with:

{\small
\begin{equation}
  \hat{\boldsymbol{\beta}}^{(l,k)} = (\mathbf{X}_{\text{train}}^{(l,k)\top} \mathbf{X}_{\text{train}}^{(l,k)} + \alpha\mathbf{I})^{-1} \mathbf{X}_{\text{train}}^{(l,k)\top} \mathbf{y}_{\text{train}}
\end{equation}
}
Here, $\alpha$ controls the regularization strength, selected via grid search in a nested cross-validation framework. All data are z-scored to ensure compatibility between fMRI and neural embeddings.

\textbf{Feature Normalization and Parallelization.} 
Both $\mathbf{y}$ and $\mathbf{X}$ are standardized to zero mean and unit variance before regression to improve numerical stability. The analysis pipeline parallelizes computations across subjects, ROIs, and layers, training $\mathcal{O}(SRL)$ models, where $S$ is the number of subjects, $R$ is the number of ROIs, and $L$ is the number of layers. This ensures efficient scalability in the layer-wise evaluation of embeddings.

This approach provides layer-wise correlation metrics ($\rho_l$), which can offer insights into how neural network representations align with ROI-specific brain activity. 

\section{Experiments and Results}
Now we present our three main experiments and results to evaluate the cross-lingual semantic alignment capabilities of LLMs, their correspondence with brain activity, and their sensitivity to hemispheric differences in neural responses.

\subsection{Performance of Large Language Models}
\label{sec:performance}
To evaluate the contextual understanding capabilities of LLMs under cross-lingual settings, 
we first design a multiple-choice test based on semantic alignment between Chinese source sentences and their English translations, using the stimuli sentences of ``The Little Prince'' used for the human fMRI data. Drawing from recent work on evaluating the robustness of LLMs in multiple choice setups \cite{zheng2024large,wang2024look}, we adopt several perturbations to test the LLMs' performance. For each original Chinese sentence from the stimuli, we generate five English options: (A) correct English translation, (B) word order scrambled, (C) part-of-speech substitution, (D) sentence structure transformation, (E) information insertion/deletion. 
Each of these English sentence options is then input into the LLM to obtain their corresponding embeddings. 


To quantitatively evaluate the cross-lingual semantic understanding capability of large language models, we propose the \textbf{Cross-lingual Semantic Alignment Accuracy (CSAA)} metric. This metric is defined as the proportion of cases in which the model correctly identifies the true translation (option A) as the most semantically similar candidate to the original Chinese sentence, as measured by cosine similarity in the embedding space.

We define an indicator variable $\delta_i$ for each sample $i$ as follows:
{\small
\begin{equation} 
\delta_i = 
\begin{cases}
1, & \text{if } \operatorname{argmax}_{x} \left( \cos\left( \mathbf{v}_{c_i}, \mathbf{v}_{e_{i,x}} \right) \right) = A \\
0, & \text{otherwise}
\end{cases}
\end{equation}
}

The CSAA is then defined as:
{\small
\begin{equation}
\text{CSAA} = \frac{1}{N} \sum_{i=1}^{N} \delta_i
\end{equation}
}
where $N$ is the total number of Chinese sentences, $\mathbf{v}_{c_i}$ denotes the embedding of the $i$-th Chinese sentence, and $\mathbf{v}_{e_{i,x}}$ denotes the embedding of the $x$-th English candidate for the $i$-th sentence. A higher CSAA indicates stronger cross-lingual semantic alignment ability of the model.

\begin{figure}[htbp]
    \centering
    \includegraphics[width=0.85\linewidth]{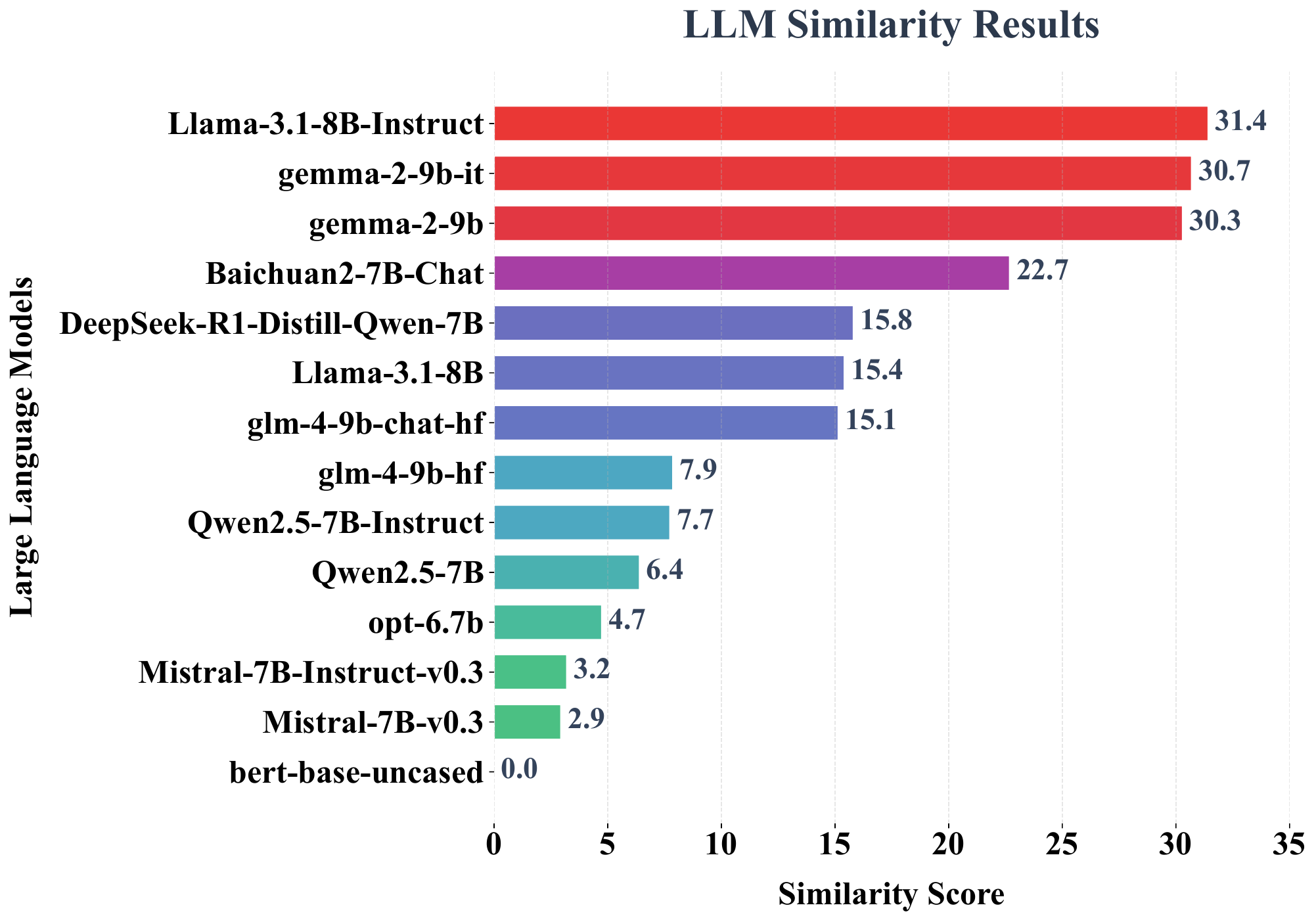}
    \caption{Performance comparison of 14 LLMs}
    \label{fig:llm}
\end{figure}

The results in Figure~\ref{fig:llm} show that the Llama-3.1-Instruct version leads with 31.4 points, followed by two Gemma variants (30.7 and 30.3). A sharp performance drop occurs after the top three models, with mid-tier scores (22.7-15.1) dominated by Baichuan2 and DeepSeek variants, while glm-4.9b (7.9) and Qwen2.5-7B (6.4) anchor the lower end. The red-to-green gradient visually reinforces score disparities, highlighting a \textgreater28-point gap between best and worst performers. Consistently, we notice that the instruction-tuned models always perform better than the base models in each model family. 

This experiment demonstrates that LLMs exhibit varying degrees of cross-lingual semantic alignment capabilities. Instruction tuning consistently improves model performance, showing the importance of fine-tuning and alignment strategies in enhancing semantic understanding. These findings establish the foundation for further investigations into how semantic alignment translates into neural representations.

\begin{figure*}[htbp]
    \centering
    \includegraphics[width=0.9\textwidth]{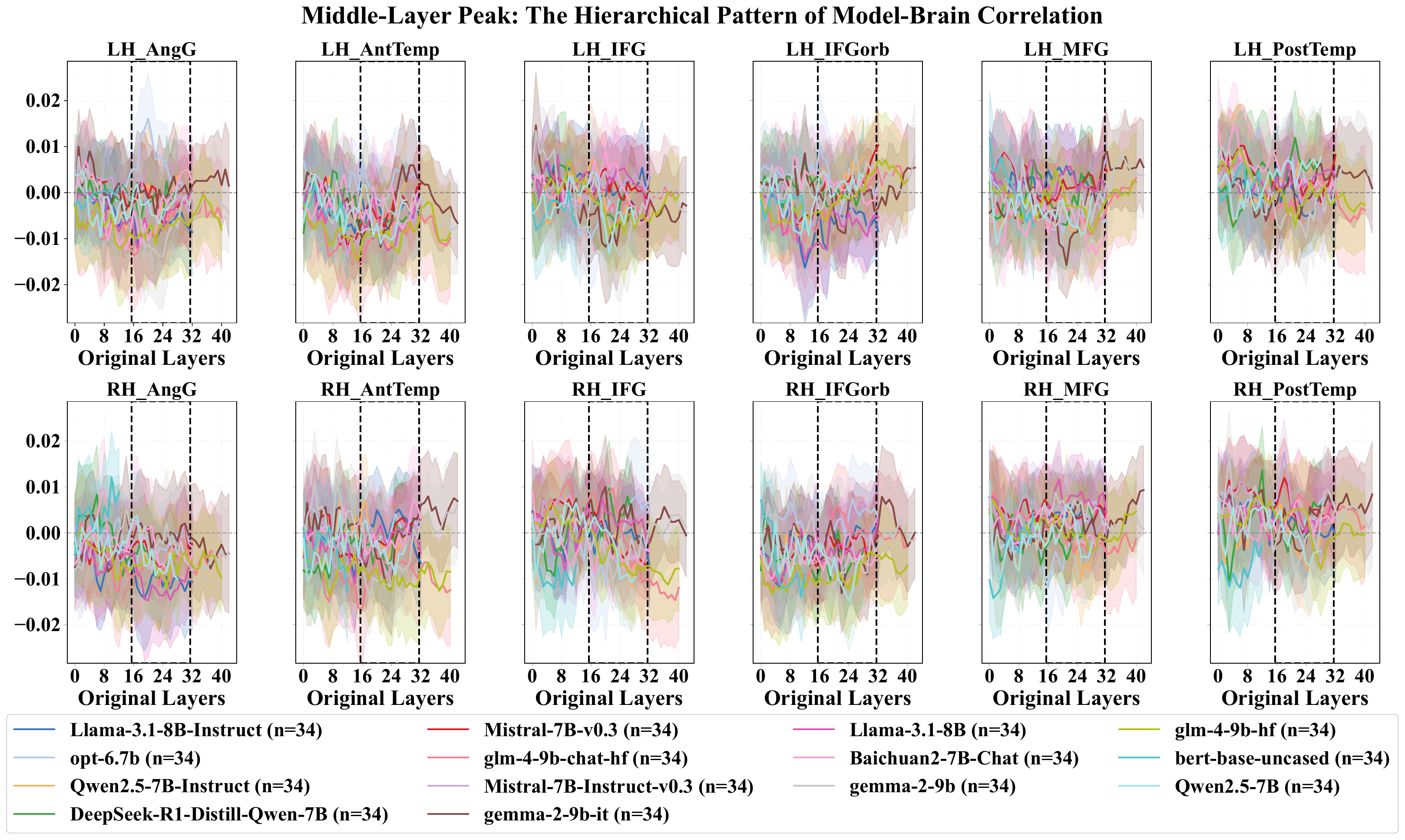}
    \caption{Correlation between model predictions and brain activity across layers. Shaded areas are 95\% confidence intervals. }
    \label{fig:all_ROIs_comparison}

\end{figure*}

\subsection{Model Performance and Activation Correlation}
\label{sec:performance_corr}
Next, to comprehensively assess the correspondence between models and brain activity, we compared model performance across multiple brain regions and layers.


The comparative evaluation of various computational models across all brain regions is summarized in 
Figure \ref{fig:all_ROIs_comparison}, 
which displays the layer-wise correlation coefficients together with their 95\% confidence intervals. Each curve represents the average performance trajectory of a given model, with shaded regions indicating confidence bounds. 
Notably, we found that in most cases, models achieve peak predictive performance at their intermediate layers, not the final layer, consistent with previous findings~\cite{mischler2024contextual}.

\begin{figure}[ht]
    \centering
    \includegraphics[width=0.9\linewidth]{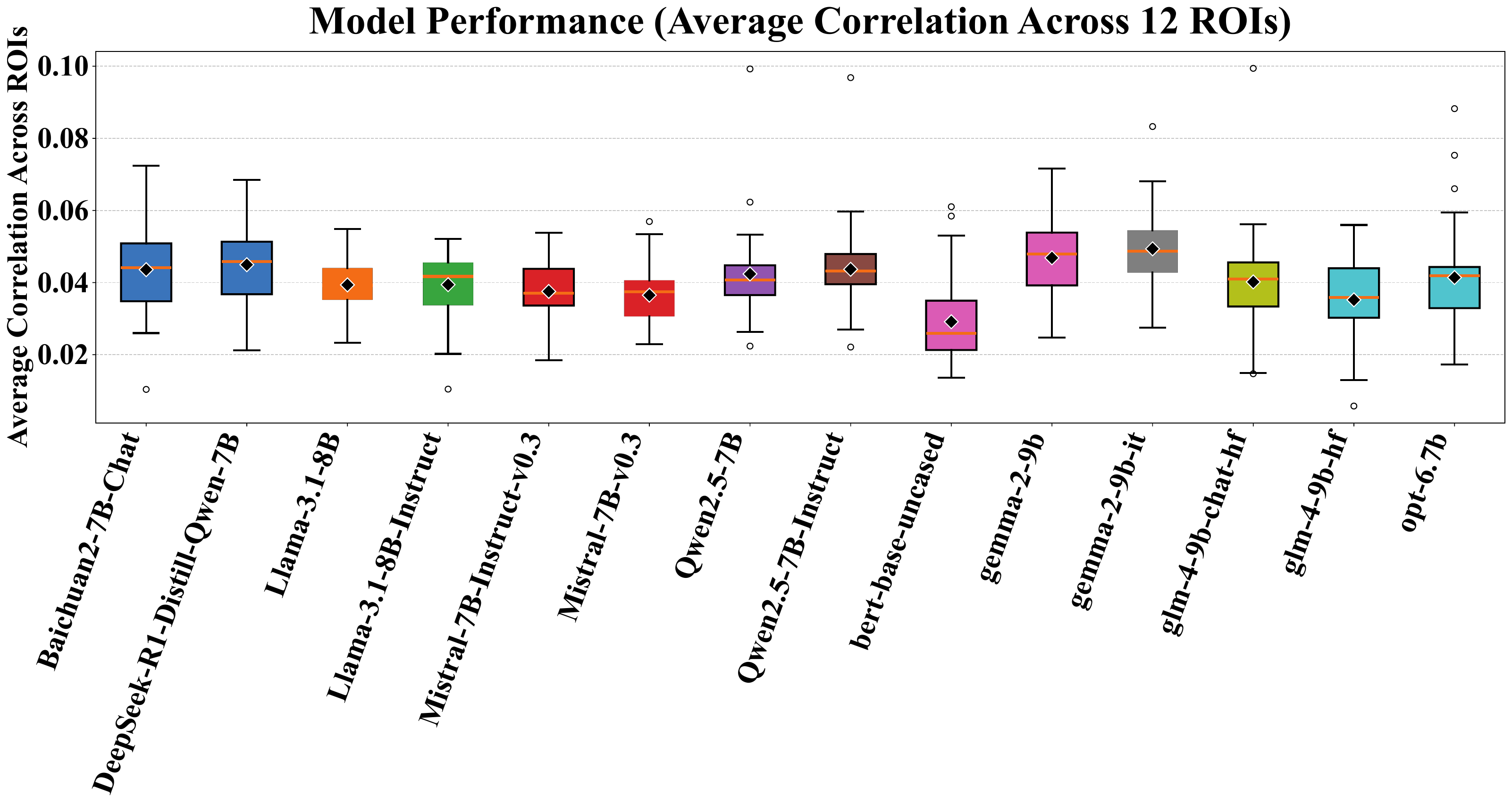}
    \caption{Average correlation of LLMs across 12 ROIs.}
    \label{fig:boxplot_model_avg_correlation}
\end{figure}

To further compare models, we selected the optimal layer exhibiting peak performance as the model's output representation and examined the average correlation across 12 ROIs, as shown in Figure~\ref{fig:boxplot_model_avg_correlation}. 
Our results show that LLMs consistently yield higher correlation metrics than the base BERT model across brain regions.

\begin{figure}[htbp]
    \centering
    \includegraphics[width=0.97\linewidth]{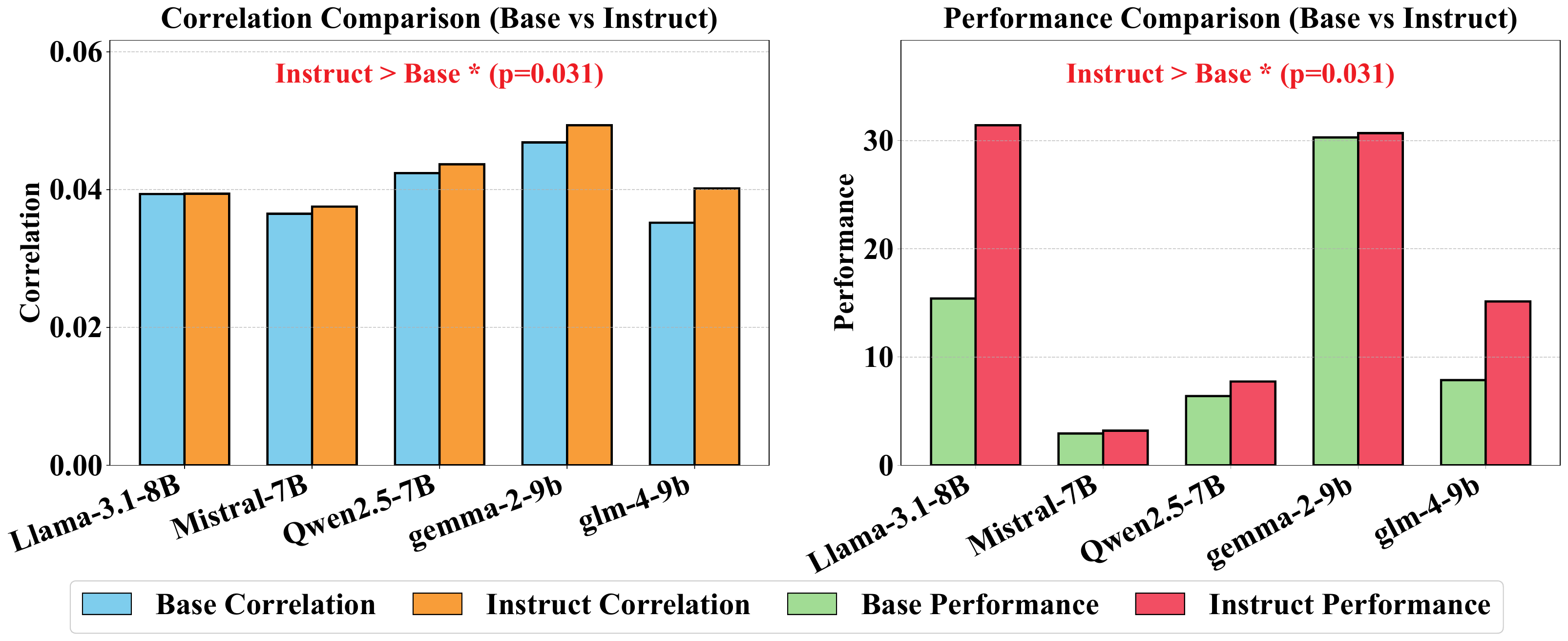}
    \caption{Comparison of instruction-tuned versus base models across performance and activation correlations}
    \label{fig:Instruct}
\end{figure}

From the 14 LLMs, we selected 5 instruction-tuned models (which have their corresponding base versions) along with their 5 corresponding base versions for comparative analysis. The selection was motivated by the intent to directly compare the impact of instruction tuning. Figure \ref{fig:Instruct}(a) presents the comparison of activation correlations, while Figure \ref{fig:Instruct}(b) shows the performance differences between base and instruction-tuned versions. 
We notice that instruction-tuned models exhibit performance improvements in both  Correlation and Performance metrics when compared with the base models. Specifically, the p-value for the Correlation Change indicates a trend toward significance, while the Performance Change (permutation test, $p = 0.03125$) demonstrates that the observed performance gains are not only substantial in magnitude but also consistent enough to reach statistical significance. These results underscore the potential of instruction-tuning in enhancing model behavior.

\begin{figure}[htbp]
    \centering
    \includegraphics[width=0.97\linewidth]{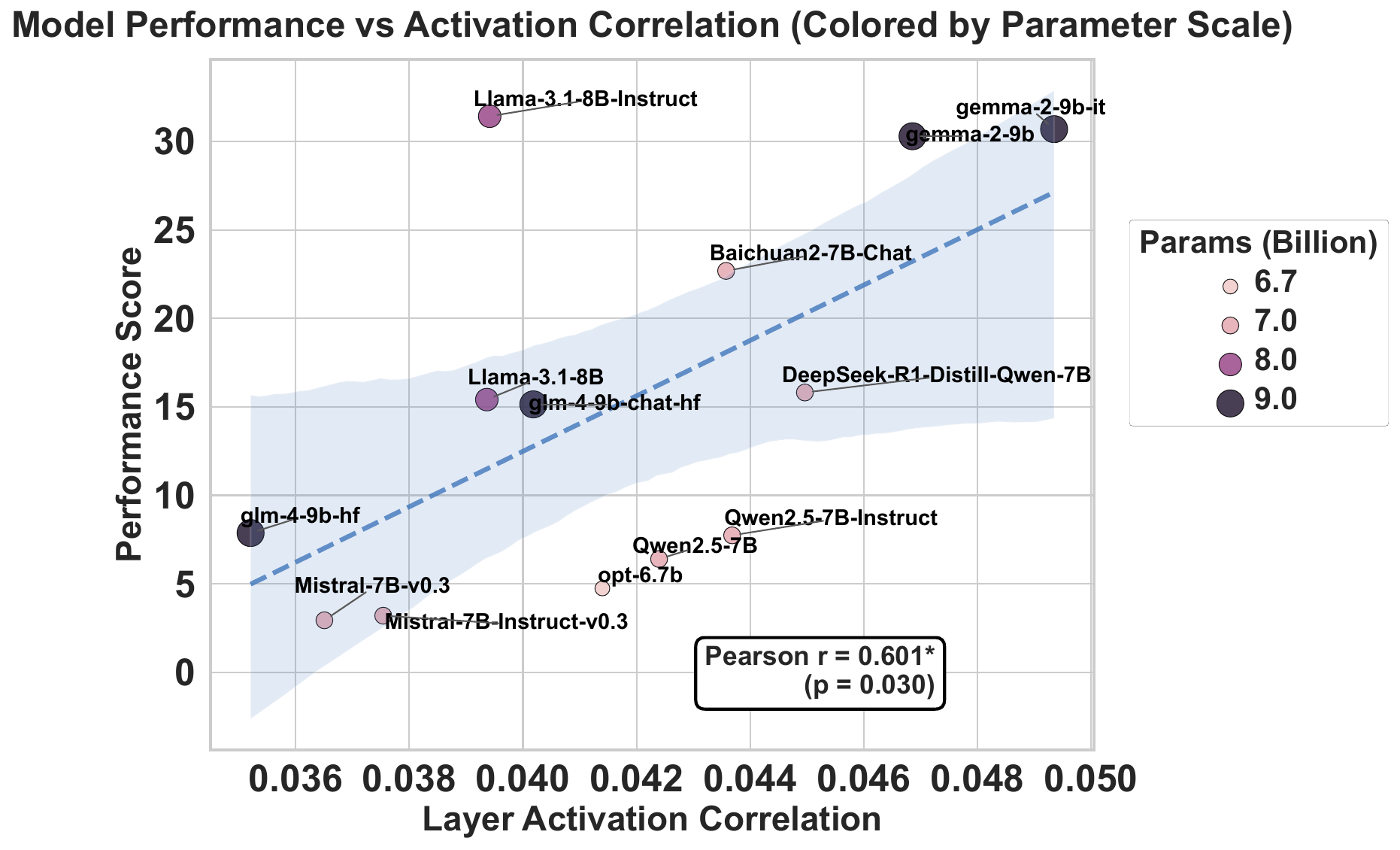}
    \caption{Correlation between model performance and activation patterns across LLMs}
    \label{fig:Performance_Activation}
\end{figure}

We then present the relationship between model performance and layer activation correlations in Figure \ref{fig:Performance_Activation}, where higher performance scores show weak positive correlation with activation values across 6.7B to 9B parameter models (Pearson correlation: $r = 0.601*$, $p = 0.030$). Notable examples include Llama-3-1-8B-Instruct (6.7B) and gemma-2.9b (9B), with activation patterns demonstrating parameter-scale dependent clustering (pink-to-purple gradient). The scatter plot's white grid background and annotated confidence interval (light blue shading) enhance comparative analysis of model architectures.

Taken together, these results reveal a clear relationship between model performance and their neural activation correlations across brain regions, with intermediate layers showing peak correspondence. Instruction tuning not only boosts performance but also strengthens alignment with brain activity, suggesting that model optimization can bring artificial and biological language processing closer together.

\subsection{Left-Right Hemispheric Asymmetry}
\label{sec:left-rigjt}

\begin{figure*}[htbp]
    \centering
    \includegraphics[width=0.85\textwidth]{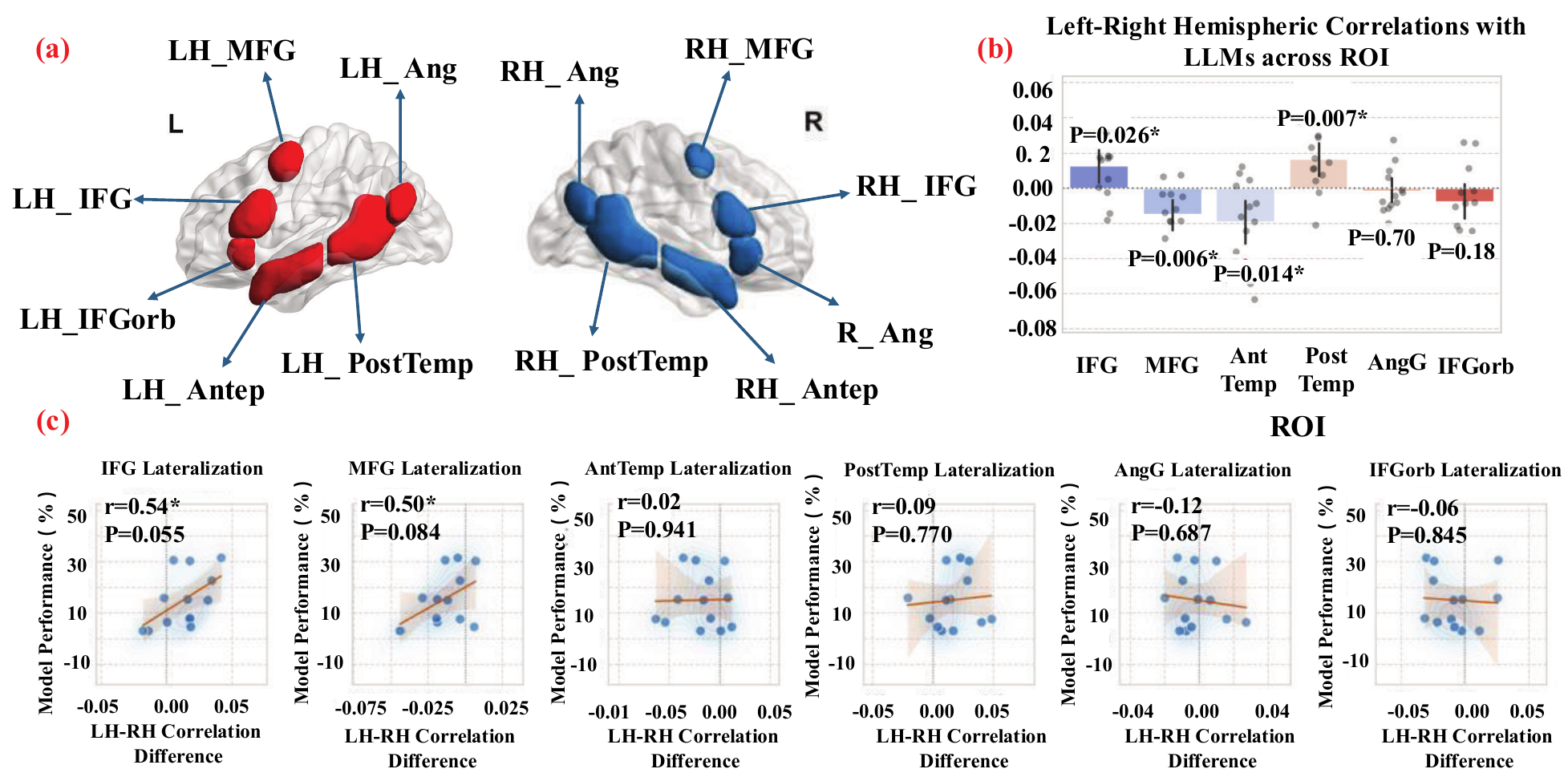}
    \caption{(a) illustrates the localization of ROIs in both hemispheres; (b) displays the left-minus-right (LH-RH) correlation differences in ROI-LLM associations; (c) examines the relationship between LH-RH asymmetry and model performance.}
    \label{fig:lr}

\end{figure*}
In the third experiment, we investigate the lateralization patterns of neural activity associated with LLM 
processing across key language-related brain regions, focusing on the asymmetry between left and right hemispheres and its relationship to model performance. In Figure \ref{fig:lr}(a), we visualize the localization of ROIs in both hemispheres.

Then, we observed left-right hemispheric asymmetry in neural activity correlations between specific brain ROIs and LLMs. In Figure \ref{fig:lr}(b), the inferior frontal gyrus (IFG) and posterior temporal (PostTemp) regions showed stronger left-hemispheric dominance in LLM-related neural correlations, consistent with their roles in core language functions like production and comprehension~\cite{hu2023precision}. Conversely, the middle frontal gyrus (MFG) and anterior temporal (AntTemp) regions exhibited stronger right-hemispheric involvement, possibly reflecting specialization in tasks such as metaphor processing, contextual integration, and cross-modal semantics. Right MFG correlations may relate to LLMs' demands for cognitive control in complex narratives~\cite{japee2015role}, while right AntTemp activity likely supports multimodal semantic representations (One-Sample t-test: IFG: $p=0.025$; PostTemp: $p=0.007$; MFG: $p=0.005$; AntTemp: $p=0.001$). These findings align with the classical language lateralization hypothesis of left-hemisphere dominance for syntax/semantics and emerging evidence of right-hemisphere contributions to cognitive control~\cite{vigneau2006meta,menon2022role}.

To examine the relationship between hemispheric asymmetry and model performance, Figure~\ref{fig:lr}(c) shows correlations between LH-RH differences and performance metrics. Among the six ROIs analyzed, IFG and MFG lateralization were potentially linked to performance, showing positive relationships (Pearson correlation: IFG: $r=0.54$, $p=0.055$; MFG: $r=0.50$, $p=0.084$). No significant correlations were found for other regions (Pearson correlation: AntTemp: $r=0.02$, $p=0.941$; PostTemp: $r=0.09$, $p=0.770$; AngG: $r=0.09$, $p=0.687$; IFGorb: $r=-0.06$, $p=0.845$). The regulatory role of the prefrontal cortex over distributed brain regions may underlie this phenomenon~\cite{badre2018frontal}, where its lateralized functional specialization could enhance top-down coordination of cognitive resources, mirroring the efficiency optimization observed in computational models.

This experiment confirms that hemispheric lateralization plays a significant role in the neural representation of language-related processes as captured by LLMs. The observed left-hemisphere dominance in core language areas and right-hemisphere contributions to higher-level cognitive functions align well with existing neurocognitive theories. 

\section{Discussion and Conclusion}

Understanding how LLMs align with human brain activity during sentence processing offers important insights into the parallels and divergences between artificial and biological language systems~\cite{mahowald2024dissociating,zhou2024divergences}. Our study contributes to this growing body of work through three core findings: (1) intermediate-layer alignment with neural activity, (2) the effects of instruction tuning and semantic comprehension on brain alignment, and (3) hemispheric asymmetries in neural–LLM correspondence. Below, we discuss each in turn.

\paragraph{Intermediate layers better align with brain activity.}
Across all evaluated models, we found that intermediate layers exhibit higher brain–model correlation than final layers, particularly across language-selective brain regions (e.g., IFG, PostTemp). This aligns with previous studies using intracranial EEG and MEG~\cite{mischler2024contextual,zhou2024divergences}, and confirms that middle-layer representations in LLMs carry rich linguistic information that most closely mirrors human sentence-level processing.

Our use of fMRI complements these prior modalities by providing broader spatial coverage across regions and hemispheres. These results emphasize the importance of considering layer-wise architecture when interpreting the cognitive fidelity of LLMs, reinforcing the notion that later layers, often optimized for downstream tasks, may abstract away from biologically grounded representations.

\paragraph{Instruction tuning enhances brain alignment and comprehension.}
We compared instruction-tuned models with their base versions across both semantic performance and brain activation correlation. Instruction-tuned variants consistently outperformed the base models, with improvements observed in both CSAA metric and neural alignment measures (Figure~\ref{fig:Instruct}). These gains were statistically significant in performance 
and showed marginal significance in brain correlation, echoing trends reported in~\citet{ren2024large}.

Importantly, we shift the interpretive focus away from model size, a dominant narrative in prior work~\cite{bonnasse2024fmri}, toward semantic comprehension ability as a more meaningful predictor of neural alignment. Our results reveal a scaling law: within the 6.7B–9B parameter range, models with stronger comprehension capabilities (higher CSAA scores) also exhibit significantly higher similarity to human brain activation patterns (Pearson $r = 0.601$, $p = 0.030$; Figure~\ref{fig:Performance_Activation}).

This suggests that optimization techniques that improve semantic understanding (such as instruction tuning) not only enhance task performance but also produce representational structures more consistent with human neural activity. Future model evaluation efforts may benefit from incorporating neural alignment as an additional axis of assessment.

\paragraph{Hemispheric asymmetry reflects functional specialization}
Our third finding addresses the hemispheric organization of LLM–brain alignment. Consistent with classical neurocognitive theories, we observed left-hemispheric dominance in the inferior frontal gyrus (IFG) and posterior temporal (PostTemp) regions, areas central to syntactic processing and semantic integration~\cite{hu2023precision,vigneau2006meta}. Conversely, the right hemisphere showed stronger correlations in the middle frontal gyrus (MFG) and anterior temporal (AntTemp), suggesting involvement in metaphor, cross-modal semantics, and cognitive control~\cite{japee2015role,menon2022role}.

These asymmetries were not only statistically robust (e.g., IFG: $p=0.025$, AntTemp: $p=0.001$; Figure~\ref{fig:lr}b), but also showed positive trends with model performance in specific ROIs. Notably, greater left–right differences in IFG and MFG were associated with higher CSAA scores (Pearson $r=0.54$ and $0.50$, respectively). This suggests that lateralized neural specialization, particularly in prefrontal regions, may support more efficient language representations, paralleling the efficiency gains observed in well-optimized LLMs~\cite{badre2018frontal}.

Interestingly, other regions (e.g., angular gyrus, IFGorb) showed no significant lateralization–performance correlation, possibly reflecting their roles in bilateral, domain-general processes such as semantic integration or the default mode network~\cite{kuhnke2023role}.

\paragraph{Implications and Future Directions.} Together, our findings paint a nuanced picture of how current LLMs reflect, and diverge from, human language systems. Instruction tuning and semantic comprehension capacity, not merely parameter size, are key drivers of neural similarity. Moreover, the correspondence is anatomically and functionally specific: intermediate-layer embeddings align best with neural signals, and this alignment respects known hemispheric specializations in the brain.

These insights raise exciting opportunities for future research. For instance, should training procedures explicitly target alignment with brain data to produce more cognitively plausible models? Could hybrid neuro-symbolic LLMs better reflect distributed and lateralized processing patterns observed in the brain? Our results lay the groundwork for these explorations by bridging computational performance with neurobiological plausibility in future work.

\newpage
\bibliography{aaai2026}

\end{document}